\documentclass[sigconf]{acmart}

\AtBeginDocument{%
  }
\usepackage[table]{xcolor}

\usepackage{algorithm}
\usepackage{algpseudocode}
\usepackage{amsmath}
\usepackage{array}
\usepackage{balance}
\usepackage{booktabs}



\usepackage{amssymb}      
\usepackage{bm}           
\usepackage{cleveref}    
\setcopyright{acmlicensed}
\copyrightyear{2025}
\acmYear{2025}
\acmDOI{10.1145/3746027.3755469}
\acmConference[MM '25]{MM '25, October 27--31, 2025, Dublin, Ireland}{October 27--31,
  2025}{Dublin, Ireland}
\acmISBN{979-8-4007-2035-2/2025/10}

\begin{document}

\title{Dynamic Residual Encoding with Slide-Level Contrastive Learning for End-to-End Whole Slide Image Representation}


\author{Jing Jin$^*$ }
\affiliation{%
  \institution{School of Computer Science and Enginecring, Central South University}
  \city{Changsha}
  \country{China}}
\email{jingjin@csu.edu.cn}

\author{Xu Liu$^*$}
\affiliation{%
  \institution{School of Computer Science and Enginecring, Central South University}
  \city{Changsha}
  \country{China}}
\email{lx67l1@csu.edu.cn}

\author{Te Gao}
\affiliation{%
  \institution{School of Computer Science and Enginecring, Central South University}
  \city{Changsha}
  \country{China}}
\email{gaote1021@csu.edu.cn}

\author{Zhihong Shi}
\affiliation{%
  \institution{School of Computer Science and Enginecring, Central South University}
  \city{Changsha}
  \country{China}}
\email{shizhihong@csu.edu.cn}

\author{Yixiong Liang}
\affiliation{%
  \institution{School of Computer Science and Enginecring, Central South University}
  \city{Changsha}
  \country{China}}
\email{yxliang@csu.edu.cn}

\author{Ruiqing Zheng}
\affiliation{%
  \institution{School of Computer Science and Enginecring, Central South University}
  \city{Changsha}
  \country{China}}
\email{rqzheng@csu.edu.cn}

\author{Hulin Kuang$^{\dagger}$}
\affiliation{%
  \institution{School of Computer Science and Enginecring, Central South University}
  \city{Changsha}
  \country{China}}
\email{hulinkuang@csu.edu.cn}

\author{Min Zen}
\affiliation{%
  \institution{School of Computer Science and Enginecring, Central South University}
  \city{Changsha}
  \country{China}}
\email{zengmin@csu.edu.cn}

\author{Shichao Kan$^{\dagger}$}
\affiliation{%
  \institution{School of Computer Science and Enginecring, Central South University}
  \city{Changsha}
  \country{China}}
\email{kanshichao@csu.edu.cn}

\renewcommand{\shortauthors}{Jing Jin et al.}


\renewcommand{\shortauthors}{Jing Jin et al.}




\begin{abstract}
Whole Slide Image (WSI) representation is critical for cancer subtyping, cancer recognition and mutation prediction.
Training an end-to-end WSI representation model poses significant challenges, as a standard gigapixel slide can contain tens of thousands of image tiles, making it difficult to compute gradients of all tiles in a single mini-batch due to current GPU limitations. To address this challenge, we propose a method of dynamic residual encoding with slide-level contrastive learning (DRE-SLCL) for end-to-end WSI representation. 
Our approach utilizes a memory bank to store the features of tiles across all WSIs in the dataset. 
During training, a mini-batch usually contains multiple WSIs. For each WSI in the batch, a subset of tiles is randomly sampled and their features are computed using a tile encoder. Then, additional tile features from the same WSI are selected from the memory bank. The representation of each individual WSI is generated using a residual encoding technique that incorporates both the sampled features and those retrieved from the memory bank. Finally, the slide-level contrastive loss is computed based on the representations and histopathology reports of
the WSIs within the mini-batch. Experiments conducted over cancer subtyping, cancer recognition, and mutation prediction tasks proved the effectiveness of the proposed DRE-SLCL method.
\end{abstract}

\maketitle

\let\thefootnote\relax\footnotetext{$^*$Both authors contributed equally to this research.}
\let\thefootnote\relax\footnotetext{$^{\dagger}$Corresponding authors.}

\section{Introduction}
\label{sec:intro}

\begin{figure}[t]
    \begin{center}
    \includegraphics[width=\linewidth]{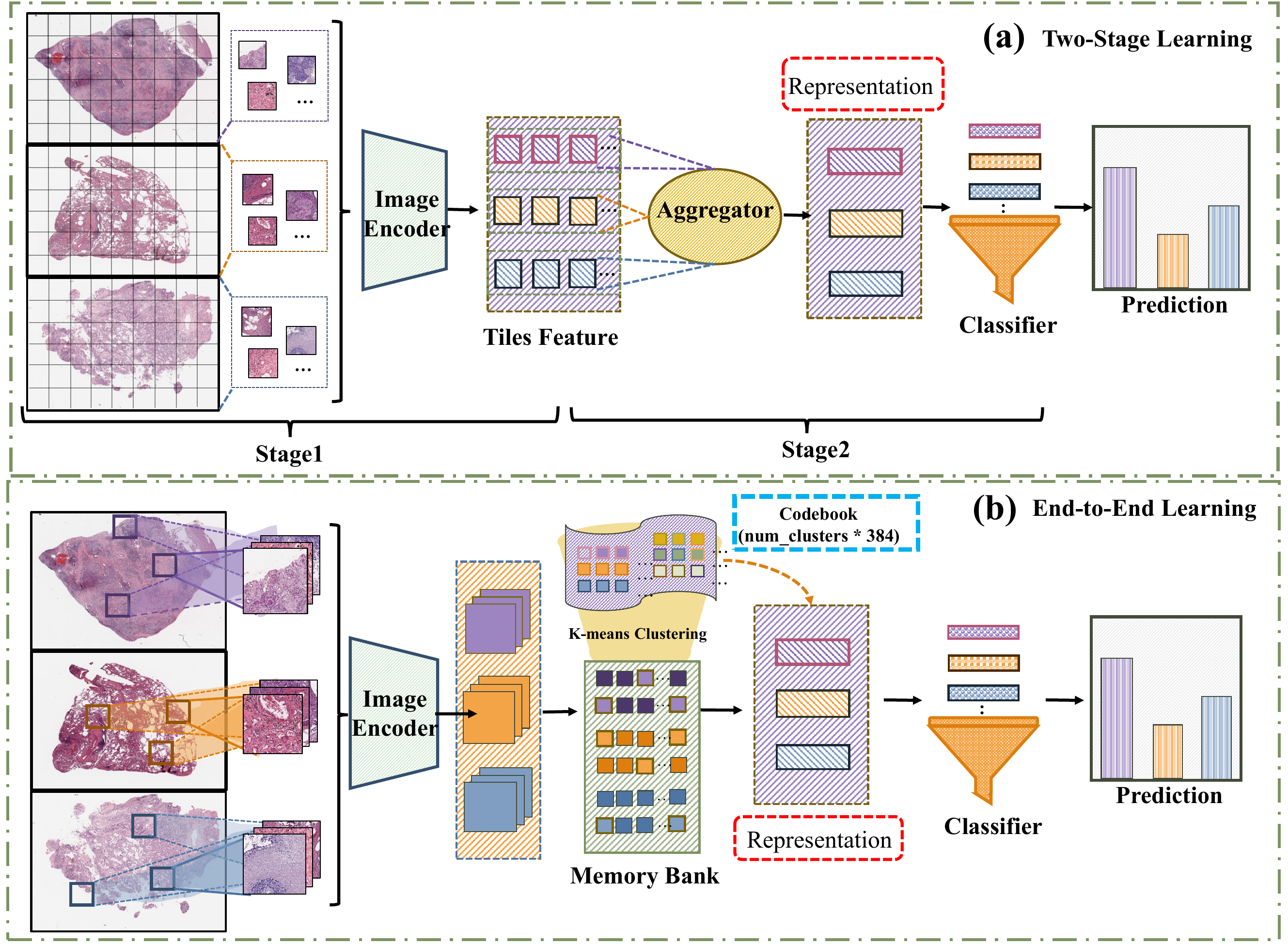}
    \end{center}
	\caption{WSI representation learning. (a) The pipeline of the two-stage learning methods and (b) the proposed end-to-end learning method.}
	\label{fig:intro}
\end{figure}

Whole Slide Image (WSI) representation is essential in computational pathology (CPath) \cite{Cpath,cpath1}, serving as the foundation for tasks such as cancer subtyping \cite{HIPT,cancer_subtyping}, cancer recognize \cite{cancer_recognize,cancer_recognize1,cancer_recognize2}, mutation prediction \cite{mutation_prediction1,mutation_prediction2}, cancer staging \cite{cancer_staging,cancer_stage1,cancer_stage2}, and prognostic assessment \cite{prognostic1,prognostic2,prognostic3}. Given that a standard gigapixel slide may consist of tens of thousands of image tiles, making it impossible for all tiles of WSIs in a batch to be updated in each epoch due to GPU limitations. Current WSI representation methods typically follow a two-stage approach, as illustrated in Figure \ref{fig:intro} (a). In the first stage, a tile representation model is trained using self-supervised techniques to capture individual tile features within the WSI. Then, these tile representations are aggregated through models like LongNet in Prov-GigaPath \cite{gigapath} or ViT in the Hierarchical Image Pyramid Transformer (HIPT) \cite{HIPT}. However, this two-stage approach may be suboptimal for WSI representation, as WSI labels are not utilized in the initial learning stage. Therefore,  we aim to develop an end-to-end learning method to enhance WSI representation by explicitly leveraging label gradients. This enables direct gradient flow between tile-level and slide-level representations, improving feature learning and overall performance.

Popular methods for WSI representation are weakly supervised and use Multiple Instance Learning (MIL) \cite{MIL1,MIL2,MIL3,dq_mil,trans_mil,ILRA-MIL}, where MIL algorithms assign slide-level labels to a set of instances and aggregate these instances through feature pooling \cite{mean-max-pooling-1,mean-max-pooling1-2}, 
Transformer \cite{trans_mil,bian2022Mixed_Supervision_Transformer,wang2023mdmil,cui2023rammil,ma2024vivo} or graph neural networks \cite{graph1,graph2,ma2024GCN-based-MIL,shu2024slidegcd,li2024knowledge,tu2025gammil}. 
Classification tasks are then performed using the aggregated features and slide-level labels. Although these methods have shown promising progress, the instance sampling process can affect model performance; specifically, the true label for a set of instances may differ from the slide-level label. 
To address this issue, we use a memory bank to store the features of all tiles, then aggregate the dynamically sampled tile features in a mini-batch with the stored tile features using a residual encoding method called the vector of locally aggregated descriptors (VLAD) \cite{vlad} during training, which offers linear computational complexity for processing thousands of tiles per WSI while effectively preserving fine-grained morphological variations through its residual-based encoding mechanism. Since the final representation is based on the features of all tiles in a WSI and the model is trained end-to-end, this approach mitigates the sampling problem and improves the performance of the WSI representation, as illustrated in Figure \ref{fig:intro} (b).

Beyond addressing computational constraints, using contrastive learning to align WSI visual features with corresponding pathology reports offers significant advantages, enhancing model generalization through cross-modal supervision. This approach has demonstrated considerable promise in natural and biomedical image processing.
In CPath, Prov-GigaPath \cite{gigapath} incorporated pathology reports to construct a contrastive loss function, optimizing the model and enhancing performance in cancer subtyping and mutation prediction. However, they didn't evaluate effectiveness of contrastive learning in an end-to-end framework. In this work, we integrate contrastive learning directly into our end-to-end training pipeline and validate the effectiveness of this approach using the lung cancer dataset from The Cancer Genome Atlas (TCGA) and the Clinical Proteomic Tumor Analysis Consortium (CPTAC). Unlike most contrastive learning methods that rely on CLIP language encoders \cite{clip}, we use the LLaMA2-7B model \cite{llama} to encode pathological reports, applying contrastive loss exclusively to slides with available diagnostic reports.

To train the model end-to-end, we implement a dynamic parameter update strategy: a mini-batch of tiles is sampled from different WSIs, and WSI features are updated as tile features are modified. The updated WSI features are then used to compute cross-entropy and contrastive losses, driving parameter updates across the model. This end-to-end approach optimizes both the feature extraction and classification components, resulting in improved overall performance. During testing, the trained image encoder extracts features for all tiles within a WSI, which are then encoded using a residual encoding method.

The contributions of this work are summarized as follows:
\begin{itemize}
    \item A method of dynamic residual encoding with slide-level contrastive learning (DRE-SLCL) is proposed for end-to-end WSI representation.
    \item We implement a dynamic parameter update strategy by incorporating a memory bank for end-to-end WSI representation learning, along with contrastive learning between the WSI and report representations directly within the end-to-end training pipeline.
    \item Our experimental results on the lung cancer dataset for cancer subtyping, cancer recognition, and mutation prediction proved the effectiveness of the proposed DRE-SLCL method.
\end{itemize}

\section{Related Work}
\label{sec:formatting}

Our work is related to multiple instance learning, contrastive learning, and existing end-to-end approaches. In the following, we review works related to them.

\subsection{Multiple Instance Learning}
Multiple Instance Learning (MIL) is a foundational approach in computational pathology, particularly for WSI analysis \cite{MIL1,MIL3,MIL2,mil+deeplearing1,mil+deeplearing2,pixel_level1,pixel_level2}, excelling when only slide-level annotations are available. WSIs are treated as "bags" containing multiple tile instances, with the model aggregating information to infer slide-level predictions.

Early MIL implementations employed simple max-pooling and average-pooling aggregation strategies for weakly supervised learning \cite{mean-max-pooling-1,mean-max-pooling1-2}. Gradually, advanced methods have introduced attention mechanisms\cite{attention-mil}. And than transformer-based MIL methods gained rapid popularity after Shao et al. introduced TransMIL \cite{trans_mil}. Following TransMIL, several key innovations emerged: Bian et al.'s approach Mixed Supervision Transformer \cite{bian2022Mixed_Supervision_Transformer}utilized hybrid supervision with limited pixel annotations; Wang et al.'s MDMIL \cite{wang2023mdmil} reduced computational demands through multiplex detection cross-attention; 
and Ma et al.'s VINO \cite{ma2024vivo} extended the paradigm to pathological video analysis.





Graph-based MIL approaches model tile relationships by representing WSIs as networks, with early GNNs \cite{graph1,graph2} treating tiles as nodes and spatial connections as edges to capture multi-level contextual information. Recent advancements have  expanded this paradigm: Ma et al.'s GCN-based MIL \cite{ma2024GCN-based-MIL} used graph convolution and attention for structural relationship analysis; Shu et al.'s SlideGCD \cite{shu2024slidegcd} pioneered inter-slide association through node classification; and Tu et al.'s GAMMIL \cite{tu2025gammil} leveraged graph attention for efficient multi-scale information fusion with reduced noise interference.

Additionally, several foundational hybrid approaches have influenced the field. CLAM \cite{CLAM} introduced a clustering-constrained attention mechanism grouping similar instances before aggregation, balancing computational demands with discriminability. HIPT \cite{CLAM} presented hierarchical learning using ViT for multi-scale processing before MIL aggregation. These earlier works remain important benchmarks due to their significant contributions.




\subsection{Contrastive Learning}
Contrastive Learning \cite{Contrastive_Learning1,Contrastive_Learning2} has become a key method in representation learning, especially for high-dimensional data such as medical and biological data \cite{medical_images1,medical_images2,medical_images3}. The core objective is to learn effective feature representations by bringing similar instances closer in latent space while pushing dissimilar instances apart. 



Early contrastive learning methods, such as SimCLR \cite{SimCLR} and MoCo \cite{MOco}, successfully learned visual representations through data augmentation, generating positive pairs while treating other samples as negatives. These approaches demonstrated that effective representations could be learned without labels by maximizing agreement between differently augmented views of the same data sample. 

Recently, contrastive learning has been employed in computational pathology to enhance WSI representations, effectively extracting key features for cancer detection and tissue segmentation \cite{WSI1,WSI2,WSI3,zhao2025colm}.
Wang et al. proposed SCL-WC \cite{wang2022scl-wc}, which used cross-slide contrastive learning to improve weakly-supervised classification through both intra-WSI and inter-WSI information exchange. CPLIP \cite{cplip} formulates a multi-image–multi-text bag-level alignment strategy for whole-slide image understanding and is primarily designed for zero-shot classification and segmentation in histopathology. Other innovations include KEP \cite{kep} that introduces a knowledge-enhanced pretraining paradigm by distilling structured medical knowledge from the PathKT into the text encoder. More recently, Yu et al. introduced CLOVER \cite{yu2025contrastive}, which integrates multi-omics data with WSI representations through contrastive learning to connect morphological and molecular tumor features.




Notably, a whole-slide pathology foundation model 
named Prov-GigaPath \cite{gigapath} integrates contrastive learning with multimodal data by aligning WSI visual features with pathology report semantics through a dual-encoder architecture greatlty improving cancer subtyping and prognostic prediction performance.



\subsection{End-to-End Training}

Despite the clear advantages of end-to-end approaches, the implementation of end-to-end models for WSI classification remains extremely limited in the literature. To our knowledge, only a handful of studies have attempted to realize this paradigm, and most of these are relatively early works with limited effectiveness. 

Chikontwe et al. \cite{chikontwe2020Chikontwe} proposed a center embedding approach that employed joint learning of instance-level and bag-level classifiers with top-k instance sampling. Takahama et al. \cite{takahama2019Takahamamulti} proposed a multi-stage approach combining patch-based classification with whole slide-scale segmentation, addressing GPU memory limitations by retaining gradient information while training the model partially. Xie et al. 
\cite{xie2020xie} introduced an end-to-end part learning approach that divided patches from a WSI into k parts based on global clustering centroids and sampled k tiles in each training epoch.  Sharma et al. \cite{sharma2021C2C}  presented the Cluster-to-Conquer framework that uses adaptive attention with KL-divergence regularization for intra-cluster attention variance. Recently, Chen et al. \cite{chen2023chen} integrated feature extraction and classification into a unified framework with attention pooling and a squared average normalization function.

\section{Methods}

    \begin{figure*}[t]
    \begin{center}
    \hspace{-9mm} 
    \includegraphics[width=1.05\linewidth]{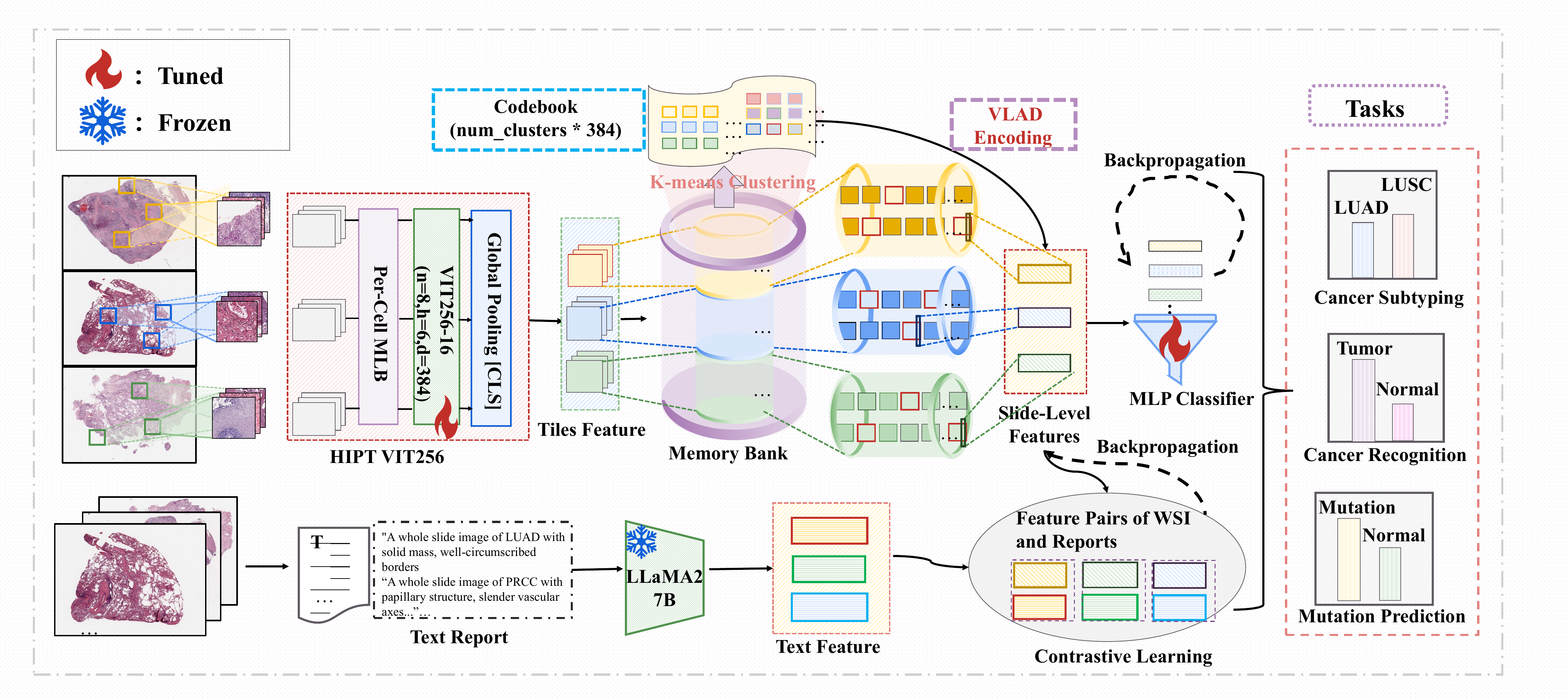}
    \end{center}
	\caption{The end-to-end whole slide image representation framework with the proposed dynamic residual encoding with slide-level contrastive learning (DRE-SLCL).}
	\label{fig:framework}
\end{figure*}

\begin{table}[htbp]
\centering
\caption{Four key components and their brief descriptions.}
\label{tab:key-components}

\begin{tabular}{p{0.14\textwidth} p{0.30\textwidth}}
\hline
\textbf{COMPONENT} & \centering\arraybackslash \textbf{BRIEF DESCRIPTION} \\
\hline
\textbf{Memory bank} & {A two-level nested dictionary that stores all tile features in the dataset, with mini-batch feature updates performed iteratively during training.} \\
\hline
\textbf{Codebook} &  {A set of prototype vectors constructed via K-means clustering on all tile features from the memory bank, kept fixed throughout all training stages.}\\
\hline
\textbf{Tile encoder}  & {A ViT-256 model, pretrained using the Hierarchical Image Pyramid Transformer (HIPT) framework, to encode tile-level features.} \\
\hline
\textbf{Report encoder } & {The LLaMA2-7B model, where we use the average of all token embeddings from the final hidden layer as the representation.}\\
\hline
\end{tabular}
\end{table}

\subsection{Method Overview}

The overall framework of our method is illustrated in Figure \ref{fig:framework}. It consists of a feature extraction network, a VLAD encoding module, and a contrastive learning module. The feature extraction network extracts tile-level features within a WSI, while the VLAD encoding module aggregates these tile features into a comprehensive WSI representation. Contrastive learning, computed with the WSI feature and the WSI report feature, aims to enhance the generalization capability of the model. For image encoding, we use the ViT-256 model pretrained on HIPT, and the WSI report encoder is the text encoder of the LLaMA2 7B model. The key components of this framework are listed in Table \ref{tab:key-components}. In the following subsections, we provide details of our method.

\begin{table}
\renewcommand{\arraystretch}{1.1}
\begin{tabular}{p{\columnwidth}}
\hline
\textbf{Algorithm 1:} \textbf{Dynamic Residual Encoding with Slide-Level Contrastive Learning (DRE-SLCL)} \\
\hline
\textbf{Input:} Whole Slide Image (WSI) $X_i = \{x_{i,1},\ldots, x_{i,N}\}$ (split into $N$ tiles), 
Pathology report $T_i$ \\
Hyperparameters: codebook size $K$, sampled tiles per WSI $r$, batch size $b$, temperature $\tau$ \\
\textbf{Output:} Slide-level prediction $\hat{Y}_i$ \\
\textbf{Workflow:} \; \\
1: \textbf{Dynamic Tile Sampling \& Feature Update} \\
\quad for each WSI $X_i$ in batch do \\
\quad\quad Randomly sample $r$ tiles: $\mathcal{S}_i \leftarrow \text{RandomSelect}(X_i,r)$ \\
\quad\quad for each tile $x_{i,j} \in \mathcal{S}_i$ do \\
\quad\quad\quad Extract feature: $f_{i,j} \leftarrow \text{ViT-256}(x_{i,j})$ \\
\quad\quad\quad Update memory bank: $\text{MemoryBank}[X_i][j] \leftarrow f_{i,j}$ \\
2: \textbf{Residual Encoding via VLAD} \\
\quad for each WSI $X_i$ in batch do \\
\quad\quad Retrieve features: $F_i \leftarrow \{\text{MemoryBank}[X_i][j] \mid j = 1,\ldots, N\}$ \\
\quad\quad if codebook $C$ uninitialized then $C \leftarrow \text{K-Means}\left(\bigcup_{i=1}^b F_i, K\right)$ \\
\quad\quad Compute residuals: \\
\quad\quad for each $c_k \in C$ do \\
\quad\quad\quad $\mathcal{X}_k \leftarrow \{f \in F_i \mid \arg\min_{c\in C}\|f - c\|_2 = k\}$ \\
\quad\quad\quad $v_k \leftarrow \sum_{f \in \mathcal{X}_k} (f - c_k)$ \\
\quad\quad Concatenate: $v_{\text{VLAD}} \leftarrow \text{Concat}([v_1, \ldots, v_K])$ \\
\quad\quad Transform: $h_i \leftarrow \text{Transformer}(v_{\text{VLAD}})$ \\
3: \textbf{Contrastive Learning Alignment} \\
\quad for each WSI $X_i$ with report $T_i$ do \\
\quad\quad Encode text: $t_i \leftarrow \text{LLaMA2-7B}(T_i)$ \\
\quad\quad Project image feature: $h'_i \leftarrow \text{Linear}(h_i)$ \\
\quad Compute logits: \\
\quad\quad $S_{\text{img2txt}} \leftarrow \tau \cdot h' \cdot t^\top$ \quad
$S_{\text{txt2img}} \leftarrow \tau \cdot t \cdot h'^\top$ \\
\quad Compute loss: \\ \quad\quad $\mathcal{L}_{\text{contrastive}} \leftarrow \frac{1}{2} \left( \text{CE}(S_{\text{img2txt}}, \mathbf{y}) + \text{CE}(S_{\text{txt2img}}, \mathbf{y}) \right)$ \\
4: \textbf{Classification \& Optimization} \\
\quad Predict label: $\hat{Y}_i \leftarrow \text{MLP}(h_i)$ \\
\quad Compute loss: $\mathcal{L}_{\text{cls}} \leftarrow \text{CE}(\hat{Y}, Y)$ \\
\quad Update parameters: $\theta \leftarrow \theta - \eta \nabla_\theta (\mathcal{L}_{\text{contrastive}} + \mathcal{L}_{\text{cls}})$ \\
\hline
\end{tabular}
\end{table}

\subsection{Dynamic Residual Encoding for WSI Representation}

Since each WSI typically contains thousands of image tiles, directly storing and computing features for all tiles would be computationally impractical. To address this, we construct a dynamic feature memory bank to manage and update features efficiently.
Specifically, we start by dividing the WSI into multiple image tiles using a fixed-size grid, denoted as $x_i \in \mathbb{R}^{256 \times 256}$, which is thus well-suited for a wide range of pathology tasks. Each tile is then processed through the VIT-256 model, pretrained within the HIPT framework, to extract its feature representation $f(x_i)$. Then, these tile features are stored in the memory bank. The memory bank is structured as a nested two-level dictionary. The keys in the first level represent unique WSI identifiers, while the keys in the second level index image tiles within each WSI. Each value stores the L2-normalized feature representation $f(x_i)$ of a tile. This hierarchical dictionary setup efficiently manages tile features, providing quick access to specific WSIs and their tile locations, thereby optimizing both storage and retrieval processes.

After extracting the tile features, we adopt the VLAD module to encode the tile features in a WSI into a WSI feature.
First, a codebook is constructed by applying $K$-means clustering to all  WSIs' tile features stored in the memory bank, as follows:
\begin{equation}
C = \{\bm{c}_1, \bm{c}_2, \dots, \bm{c}_K\}.
\end{equation}
The codebook $C$, containing $K$  cluster centers, captures  morphological patterns across diverse tissue types and remains fixed during training. This static design ensures consistent residual encoding by maintaining a stable semantic reference space, preventing semantic drift that would occur with dynamic codebook updates. 

For each WSI, all its tile features are sampled from the memory bank and quantized to their nearest codewords, denoted as $\bm{c}_{\text{nearest}}$. The residual for each feature $f(x_i)$ with respect to its assigned codeword is subsequently computed as:
\begin{equation}
    \bm{r}_i = f(x_i) - \bm{c}_{\text{nearest}}.
\end{equation}
The residual vector $\bm{r}_i$ represents the deviation of the feature $f(x_i)$ from its assigned codeword. These residuals effectively capture the subtle differences between the local features and the cluster centers. 

Next, the residual vectors of all tile features from this WSI are aggregated to obtain a consolidated residual representation relative to each codeword:
\begin{equation}
    \bm{v}_k = \sum_{x_i \in \mathcal{X}_k} \bm{r}_i, \quad k = 1, 2, \dots, K,
\end{equation}
where $\mathcal{X}_k$ represents the set of all image tiles most closely associated with codeword $\bm{c}_k$. This approach accumulates all relevant residuals for each cluster center, thus generating a global representation that captures the relationship between the local features of the entire WSI and the cluster centers.

Finally, the accumulated residuals of all codewords are concatenated to form the residual encoding representation for each WSI, as follows:
\begin{equation}
    \bm{v}_{\text{re}} = [\bm{v}_1, \bm{v}_2, \dots, \bm{v}_K] \in \mathbb{R}^{K \times d},
\end{equation}
where $d$ denotes the dimension of each feature vector. To further enhance the discriminative power of the residual encoding vector, the L2 normalization is applied to the VLAD representation.

After generating the residual encoding vector, we further enhance its expressive power by introducing a Transformer layer for feature enhancement. 

In each iteration during training, we randomly sample a batch of slides and select a subset of tiles from these slides to form a mini-batch. When tile features are updated, the corresponding residual encoding representation of the WSI is recalculated. 
We refer to this process as dynamic residual encoding.

\subsection{Slide-Level Contrastive Learning}

To improve generalization, we propose a slide-level contrastive learning method that aligns visual features with textual features. The visual features are represented by the residual encoding vector of the WSI, while the textual features are extracted from the corresponding pathological reports. 

Specifically, textual features are extracted using the LLaMA2 7B model, generating 4096-dimensional vectors that capture the semantic content related to pathological conditions, diagnostic results, and other relevant information in pathological reports. The extracted textual feature vectors are then L2-normalized to ensure consistent norms across the feature space. To align the visual features with the textual features, we apply a linear mapping to the visual features, embedding them into the same feature space as the textual features. This mapping is achieved through a linear layer, ensuring that the visual feature vectors and textual feature vectors both have the same dimensionality (4096 dimensions). This uniform feature dimensionality allows for direct comparison between the two modalities within the same feature space.

In contrastive learning loss calculation, we adopt an approach similar to CLIP, using cross-entropy loss to align visual and textual features. Specifically, the loss calculation involves two directions: visual-to-text and text-to-visual. During the forward pass, both visual and textual features are first linearly mapped and normalized. Then, the similarity between the two feature sets is measured by computing a similarity matrix (logits). Let the visual features be represented as the matrix $\mathbf{V} \in \mathbb{R}^{N \times d}$ and the textual features as the matrix $\mathbf{T} \in \mathbb{R}^{N \times d}$. The similarity matrices between them are calculated as follows:
\begin{equation}
    \bm{S}_{\text{str}} = \sigma_1 \times \mathbf{V} \mathbf{T}^T,
\end{equation}

\begin{equation}
    \bm{S}_{\text{rts}} = \sigma_2  \times \mathbf{T} \mathbf{V}^T,
\end{equation}
where $\sigma_1 $ and $\sigma_2$ are learnable scalar parameters used to adjust the magnitude of the similarity. Initially, they are set to $\exp(\log(1/0.07))$, which helps enhance the model's sensitivity to feature differences in the early stages of training. As training progresses, they are dynamically updated to ensure the quality of feature alignment.

Subsequently, the model's ability to learn cross-modal matching relationships is assessed by calculating the cross-entropy loss for both slide-to-report and report-to-slide alignments. Let the labels be $y = [0, 1, \dots, N-1]$. The cross-entropy loss for slide-to-report and report-to-slide alignment is defined as:
\begin{equation}
    \mathcal{L}_{\text{str}} = \frac{1}{N} \sum_{i=1}^N -\log \frac{\exp(\bm{S}_{\text{str}}[i, y[i]])}{\sum_{j=1}^N \exp(\bm{S}_{\text{str}}[i, j])},
\end{equation}
\begin{equation}
    \mathcal{L}_{\text{rts}} = \frac{1}{N} \sum_{i=1}^N -\log \frac{\exp(\bm{S}_{\text{rts}}[i, y[i]])}{\sum_{j=1}^N \exp(\bm{S}_{\text{rts}}[i, j])}.
\end{equation}

Finally, the contrastive loss is the average of them:
\begin{equation}
    \mathcal{L}_{\text{contrastive}} = \frac{1}{2} (\mathcal{L}_{\text{str}} + \mathcal{L}_{\text{rts}}).
\end{equation}

The contrastive learning approach ensures that visual and textual features are closely aligned in semantic space and effectively enhances the model's cross-modal retrieval capabilities and improves the representation of pathological features, capturing the intermodal relationships with higher accuracy.

\subsection{Three distinct stages of the End-to-end Strategy}

Our DRE-SLCL method mainly comprises three distinct stages: preparation, training, and testing to develop our end-to-end WSI representation framework.

\paragraph{\textbf{The preparation stage:}} We process all raw WSI data from the datasets by segmenting each image into non-overlapping tiles of size 256×256 pixels. Then we utilize the tile encoder to extract initial features from all tiles, which are subsequently stored in the memory bank. K-means clustering  is then applied to the extracted tile features to generate a codebook that captures the dominant patterns within the feature space.

\paragraph{\textbf{The training stage:}}
In each iteration, from each WSI in a batch of 64, we randomly sample 10 image tiles - an optimal quantity determined by our ablation study in Section 4.5. 
This random sampling strategy serves as data augmentation while reducing computational overhead, ensuring diverse tile combinations across training iterations.
As the parameter of tile encoder is updated in each epoch, the tile features of the mini-batch are also updated. We use the updated tile features to replace the old tile features stored in the memory bank. Due to its nested two-level dictionary design, we can quickly locate which WSIs are updated for those tiles and complete this update process in less time. Then, we used both the updated tiles and other original tiles from the memory bank for the final WSI representation using the VLAD method.
Subsequently, an updated representation is then used for classification and to compute the contrastive loss.

In particular, to enable effective end-to-end training, we adopt a two-stage training strategy. In the first stage, we freeze the parameters of the VIT-256 model and train only the classifier, stabilizing the process and accelerating convergence. Once this baseline is set, we gradually unfreeze the VIT-256 model parameters, enabling joint fine-tuning of both the feature extractor and classifier. This staged unfreezing and end-to-end training optimizes the synergy between feature extraction and classification, leading to more refined feature representations and improved classification accuracy.

\paragraph{\textbf{The testing stage:}} The trained feature extraction model and codebook are used to produce an end-to-end representation for each WSI. The feature extractor segments the WSI into tiles, extracting features $f(x_i)$ for each tile. And then these features are going to  a robust global representation of the WSI with the VLAD module and a transformer layer. This global representation is subsequently processed by a classifier to produce the final classification result.









\section{Experiments}

In this section, we demonstrate the effectiveness of the proposed DRE-SLCL method on the TCGA and CPTAC datasets. The experiments focus on three primary tasks: (1) TCGA\_LUNG and CPTAC\_LUNG subtype classification, (2) binary classification for the presence of cancer in TCGA\_LUAD, TCGA\_LUSC, CPTAC\_CCRCC and CPTAC\_PDA, and (3) prediction of four major gene mutation types in TCGA\_LUAD. Additionally, we conduct an ablation study to assess the contribution of each module and evaluate the generalization ability of the proposed method. The following paragraphs provide detailed information on the experimental datasets, evaluation metrics, and parameter settings.

\paragraph{\textbf{Datasets and Evaluation Metrics}}

We evaluated the proposed DRE-SLCL method on two publicly available lung cancer datasets from the TCGA project: LUAD (Lung Adenocarcinoma) and LUSC (Lung Squamous Cell Carcinoma) and four cancer datasets from the CPTAC project: LUAD (Lung Adenocarcinoma), LSCC (Lung Squamous Cell Carcinoma), CCRCC (Clear Cell Renal Cell Carcinoma), PDA (Pancreatic Ductal Adenocarcinoma). For the gene mutation prediction task, we focus on predicting TP53, FAT1, LPRB1, and EGFR mutations. The evaluation metrics include the area under the curve (AUC), and the weighted F1 score. 
For cancer recognition, we report the AUC score.
For cancer subtyping and the gene mutation prediction task, we report all of the aforementioned metrics.


In particular, not all WSIs have corresponding pathology reports for contrastive learning. To fully exploit the available textual information, all WSIs with pathology reports are included in the training set for contrastive learning with visual features.




\paragraph{\textbf{Experimental Setting}}

The experiments were conducted on a single Nvidia A6000 GPU, with the proposed model implemented in PyTorch. In each experiment, we utilized the ViT-256 model pre-trained in the HIPT framework to extract tile-level features. The model was trained using Adam optimizer with a learning rate of $10^{-4}$ and a weight decay of $10^{-5}$, for a total of 100 epochs.



The training batch size was set to 64, with testing performed using batch size 1. We randomly sampled 10 tiles per WSI in each iteration, and set the codebook size k to 64 clusters. These parameters were optimized through ablation studies (Section 4.5) and all the  implementation details follow the three-stage strategy described in Section 3.4.





\subsection{Cancer Subtyping on the TCGA\_LUNG and the CPTAC\_LUNG Datasets}

In this task, we performed the binary classification of LUAD and LUSC using TCGA\_lung and CPTAC\_lung datasets. Compared to traditional block-based classification methods, the proposed approach achieves significant improvements for the weighted F1 score, as shown in Table \ref{tab:cancer_subtyping}.
Specifically, our model improved by approximately 4.83\% for the weighted F1 score compared to the best competing methods, demonstrating its effectiveness in comprehensively modeling global features.


\begin{table}[t]
    \centering
        \caption{The comparison of AUC (\%) and F1 scores (\%) for cancer subtyping on the TCGA and CPTAC lung datasets.}
    \label{tab:cancer_subtyping}
    
    \vspace{-3mm}
    \normalsize 
    \resizebox{0.47\textwidth}{!}{
    \begin{tabular}{l|c c|c c}
        \toprule
        \textbf{Methods} & \multicolumn{2}{c|}{\textbf{TCGA\_lung}} & \multicolumn{2}{c}{\textbf{CPTAC\_lung}} \\& AUC & F1-score & AUC & F1-score \\
        \midrule
        CLAM-SB\cite{CLAM} &\textbf{85.59} & 76.05 & 78.99 & 68.65 \\
        CLAM-MB\cite{CLAM} &83.29 & 75.78 & 70.27 & 57.50\\
        

        ABMIL\cite{attention-mil} &81.26 &73.98  &76.35 &64.46 \\
        MIL\cite{CLAM} &75.55 & 35.85 & 57.02 &47.12 \\
        TransMIL\cite{trans_mil} &77.62 &66.15 &75.95 &62.32 \\
        HIPT\cite{HIPT} &83.51 & 75.75 & 78.35 &69.28 \\
        Prov-GigaPath\cite{gigapath} &81.32 & 72.45 & 75.75 &63.57 \\
        C2C \cite{sharma2021C2C}   &78.56 &70.49 &75.53 &66.83
        \\
        DGR-MIL\cite{dgr-mil} &84.06 &77.43  &78.76 &71.24 \\

        \midrule
        DRE-SLCL(w/o cliploss) &83.12&79.63&78.89&74.35\\
        ABMIL\cite{attention-mil}+cliploss &84.48&75.56&77.43&67.29\\
        \rowcolor{green!15}
        \textbf{DRE-SLCL(ours)} &83.76 & \textbf{80.88} & \textbf{80.28} & \textbf{76.43}\\
        \bottomrule
    \end{tabular}
    }

\end{table}


\begin{table}[h]
    \centering
        \caption{The comparison of AUC (\%) scores for cancer recognition on the TCGA\_LUAD,TCGA\_LUSC,CPTAC\_CCRCC and CPTAC\_PDA datasets.}
    \label{tab:cancer_recognition}
    \vspace{-3mm}
    \normalsize 
     \resizebox{0.48\textwidth}{!}{
    \begin{tabular}{l|c c|c c}
        \toprule
        \textbf{Methods}&\multicolumn{2}{c|}{\textbf{TCGA dataset}}& \multicolumn{2}{c}{\textbf{CPTAC dataset}}\\
        & LUAD &
        LUSC & CCRCC & PDA \\
        \midrule
       CLAM-SB\cite{CLAM} &99.00 &99.23 & 99.28 & 83.71 \\
       CLAM-MB\cite{CLAM} &98.94 &98.66 & 98.75 & 86.32 \\
       
       
       ABMIL\cite{attention-mil} &97.73 & 98.79 &98.54 &85.58 \\
       MIL\cite{CLAM}     &98.13 &98.55 & 98.04 & 82.81 \\
       TransMIL\cite{trans_mil} &98.41 &97.09   &99.23  &82.74 \\
        HIPT \cite{HIPT}   &98.08 &99.28& 99.04 & 85.45\\
        Prov-GigaPath\cite{gigapath} &97.07 & 96.81 & 98.23 & 82.45 \\
        C2C \cite{sharma2021C2C}   &97.22 &98.98 &98.87 &89.51 \\

        DGR-MIL\cite{dgr-mil} &98.89 &98.76  &99.12 &87.63 \\
        
    
        \midrule
        DRE-SLCL(w/o cliploss) &99.02&98.43&98.75&90.87\\
        ABMIL\cite{attention-mil}+cliploss &98.29&98.57&99.05&86.23\\
        \rowcolor{green!15}
       \textbf{DRE-SLCL(ours)} &\textbf{99.26} &\textbf{99.58}& \textbf{99.53} & \textbf{92.34}\\
        \bottomrule
    \end{tabular}
    }

\end{table}


\begin{table*}[t]
    \centering
     \caption{The comparison of gene mutation scores (\%) on the LUAD dataset.}
    \label{tab:comparison_genes}
    
    \small
    \resizebox{0.8\textwidth}{!}{
    \begin{tabular}{l|cc|cc|cc|cc}
        \toprule
        \textbf{Methods} & \multicolumn{2}{c|}{\textbf{TP53}} & \multicolumn{2}{c|}{\textbf{ZGPR}} & \multicolumn{2}{c|}{\textbf{LPRB1}} & \multicolumn{2}{c}{\textbf{FAT1}} \\
        & AUC & F1-score & AUC & F1-score & AUC & F1-score  & AUC & F1-score \\
        \midrule
        CLAM-SB\cite{CLAM} &61.41 &79.10 &56.73 &70.11 &57.46 &58.25 &49.52 &16.39 \\
        CLAM-MB\cite{CLAM} &65.33 &76.19 &57.55 &63.20 &59.47 &52.53 &49.75 &47.76 \\
        

        ABMIL\cite{attention-mil}    
        &64.17 &75.47 &56.48 &66.37 &57.25 &50.25 &50.17 &45.56 \\
        MIL\cite{CLAM}     &57.36 &65.34 &56.34 &19.51 &51.28 &52.40 &45.29 &0 \\
        TransMIL\cite{trans_mil} &60.85 &78.18 &54.09  &56.52 &50.39 &51.71 &53.40 &53.40  \\
        HIPT\cite{HIPT}    &67.61 &81.88 &61.15 &69.92 &57.27&48.94 &50.56 &19.83 \\
        Prov-GigaPath\cite{gigapath} &61.83 &65.46 &56.30 &63.85 &52.88 &52.53 &50.67 &51.01 \\
        C2C\cite{sharma2021C2C}    &69.23 &79.40 &52.74 &80.00 &55.54 &50.53 &50.68 &0 \\
        DGR-MIL\cite{dgr-mil}    
        &69.12 &78.89 &58.11 &78.43 &59.36 &52.49 &52.38 &50.69 \\
        
        
        \midrule
        DRE-SLCL(w/o cliploss) &68.74&79.43&56.21&73.51&58.76&49.63&53.41&53.86\\
        ABMIL\cite{attention-mil}+cliploss &65.82&77.24&55.63&67.56&52.23&53.34&52.17&48.79\\
    \rowcolor{green!15}
    \textbf{DRE-SLCL(ours)}  &\textbf{71.33} &\textbf{82.76} &\textbf{57.13} &\textbf{80.80} &\textbf{60.17} &\textbf{50.67} &\textbf{54.94} &\textbf{54.17} \\
        \bottomrule
    \end{tabular}    }
   
\end{table*}



\subsection{Cancer Recognition on TCGA and CPTAC Datasets}

For the binary cancer recognition task on the LUAD, LUSC, CCRCC, and PDA datasets, we used the same architecture used in the previous subsection to encode tile features and generate global representations. The experimental results are shown in Table \ref{tab:cancer_recognition}. We can see that our method outperforms mainstream approaches for the AUC score, with particular robustness in distinguishing edge cases. Using dynamic residual encoding, the model dynamically adjusts the importance of different image tiles during training, leading to more accurate detection of cancer status.



\subsection{Gene Mutation Prediction on The LUAD Dataset}

To further validate the effectiveness of our model, we performed classification predictions for four key gene mutation types on the LUAD dataset. These mutations are commonly observed in lung adenocarcinoma, and the goal was to predict their presence. The results are shown in Table \ref{tab:comparison_genes}. The proposed method demonstrated superior performance in terms of AUC and F1 scores, with significant improvements over other approaches, particularly excelling in handling small sample sizes and addressing label imbalance.

\subsection{Comparison with State-of-the-Art Methods}

The proposed method was systematically compared with several state-of-the-art approaches, including CLAM-SB \cite{CLAM}, CLAM-MB \cite{CLAM}, ABMIL \cite{attention-mil}, TansMIL \cite{trans_mil}, HIPT \cite{HIPT}, MIL \cite{CLAM}, Prov-GigaPath \cite{gigapath}, DGR-MIL\cite{dgr-mil} and one of end-to-end WSI classification methods C2C \cite{sharma2021C2C}. 

    
    
    
    
    

The comparison results are summarized in Tables \ref{tab:cancer_subtyping}, \ref{tab:cancer_recognition}, and \ref{tab:comparison_genes}, which present performance across three major tasks: cancer subtyping, cancer presence classification, and mutation prediction in LUAD. We can see that the proposed DRE-SLCL method outperforms traditional MIL methods. DRE-SLCL excels at capturing interactions between image tiles and generating a comprehensive global feature representation. In both cancer recognition and mutation prediction tasks, DRE-SLCL shows superior performance and effectively handles complex biomedical imaging data. Notably, in the LUAD gene mutation prediction task, DRE-SLCL exhibits strong robustness in addressing data imbalance and small sample sizes. The improvements observed in AUC scores and F1-scores validate the effectiveness of our approach over other state-of-the-art methods.

\subsection{Analysis of Contrastive Loss Contribution}

To evaluate the contribution of contrastive loss in our model, we conducted additional experiments as shown in Tables \ref{tab:cancer_subtyping}, \ref{tab:cancer_recognition}, and \ref{tab:comparison_genes}. We implemented a version without contrastive loss (DRE-SLCL w/o cliploss), which still performed well across all tasks, outperforming most baseline methods. For TP53 prediction, this version achieved 68.74\% AUC and 79.43\% F1-score, while the complete DRE-SLCL showed further improvements (71.33\% AUC, 82.76\% F1-score). To verify that performance improvements weren't simply due to contrastive loss, we integrated it into ABMIL \cite{attention-mil} (ABMIL 
\cite{attention-mil}+cliploss). And the results remained below our method. These experiments demonstrate that DRE-SLCL's superior performance stems from the synergistic combination of our dynamic residual encoding architecture and contrastive learning strategy, rather than solely from the contrastive loss component. This finding was consistent across all three experimental tasks.

\begin{figure*}[t]
    \begin{center}
        \hspace{-3mm} 
        \includegraphics[width=\linewidth]{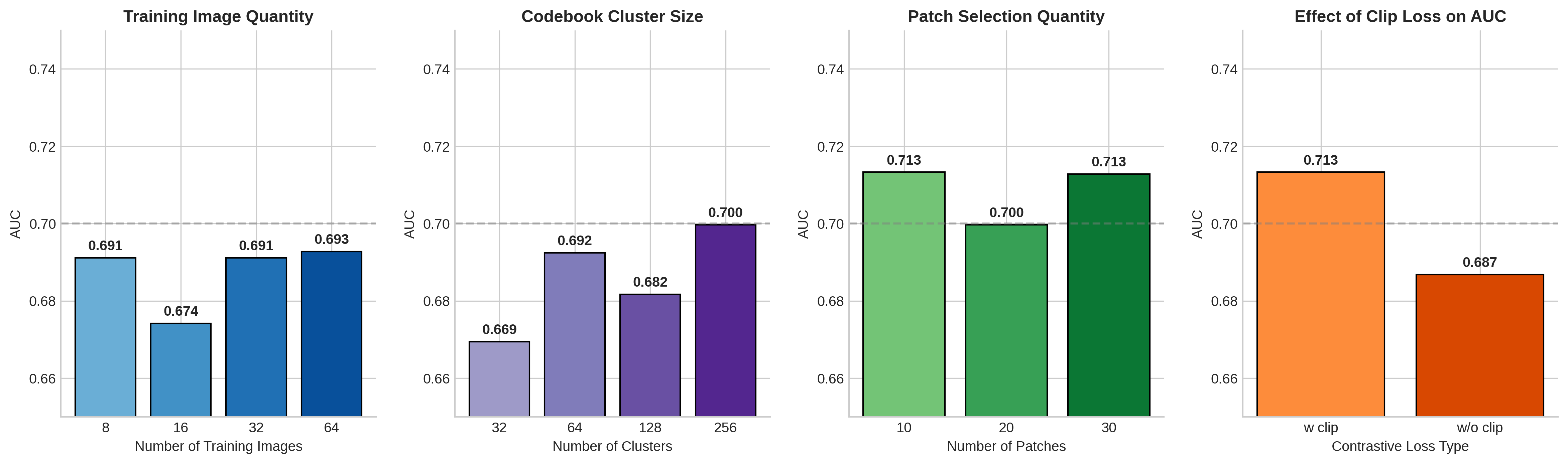}
    \end{center}
    \caption{Ablation results for TP53 gene mutation prediction.}
    \label{fig:ablation}
\end{figure*}


\subsection{Analysis of Computational Efficiency and Model Complexity}
The proposed framework demonstrates computational efficiency through a lightweight architectural design. The core backbone employs a pretrained ViT-256 comprising approximately 22M parameters, while the  VLAD encoding module and Transformer aggregator contribute fewer than 5M parameters, a total model size of under 27M parameters—significantly smaller than recent foundation model-based approaches such as Prov-GigaPath (1.1B parameters). 

The dynamic tile sampling mechanism enables training on a representative subset of tiles per WSI (r=10) rather than processing all tiles in each epoch, substantially reducing computational burden and allowing single training epoch completion in approximately 5–10 minutes on a single NVIDIA A6000 GPU. During inference, while all tiles are processed to ensure complete information utilization (typically requiring several minuites per WSI), the end-to-end design eliminates redundant I/O and preprocessing steps compared to traditional two-stage pipelines, and the memory bank stores intermediate embeddings in CPU memory without consuming GPU resources. This approach achieves an optimal balance between model capacity and computational cost, making it well-suited for scalable deployment in resource-constrained clinical environments.

\subsection{Ablation Study}
To evaluate the contribution of different components of our method, we conducted an ablation study using the gene mutation prediction task for TP53. The study focused on several aspects: codebook size, batch size, number of tiles selected per WSI, inclusion of contrastive loss, and whether the model was trained end-to-end. The results are visualized in Figure 3.

\paragraph{\textbf{Effect of the batch size:}} 
The impact of different batch sizes (8, 16, 32, 64) on AUC is presented in the first subfigure of \cref {fig:ablation}. The batch size was chosen based on contrastive learning theory (larger batches provide more negative samples),  dataset scale, and hardware constraints (NVIDIA A6000 with 48GB VRAM). And the result shows batch size 64 achieves a balance between computational feasibility and model performance with the highest AUC. 

\paragraph{\textbf{Effect of codebook size:}} 
The effect of varying the number of clusters in the codebook (32, 64, 128, 256) is shown in the second subfigure. The choice follows the classical VLAD encoding framework where cluster centers typically range from 32 to 256. K=256 is a widely adopted setting that balances descriptive power and computational efficiency. Increasing the size of the cluster generally improves the AUC, with the best appears at 256 clusters, indicating that a larger codebook can effectively capture nuanced features.

\paragraph{\textbf{Effect of the number of tiles:}} 
The third subfigure in Figure 3 shows the effect of selecting different numbers of patches per WSI (10, 20, 30) on AUC. The tile sampling follows MIL principles, where sampling a representative subset (r=10) reduces computational complexity from O(N×d) to O(r×d), where N is the total number of tiles and d is the feature dimension. In MIL, not all instances contribute equally to the final prediction. In fact, excessive sampling may introduce noise or irrelevant patterns. And the results also shows that both 10 and 30 patches better performance compared to 20.

Based on the ablation study, we selected the optimal parameters: batch size of 64, codebook size of 256 clusters, 10 patches per WSI, and inclusion of contrastive loss. This combination consistently provided the best overall performance, demonstrating its effectiveness in capturing the complex features of the whole slide images.

\section{Conclusion}

In this work, we presented a novel method called Dynamic Residual Encoding with Slide-Level Contrastive Learning (DRE-SLCL) for end-to-end Whole Slide Image (WSI) representation. Our approach integrates dynamic residual encoding and slide-level contrastive learning into a unified framework, effectively addressing the limitations of traditional two-stage learning processes. By incorporating a memory bank and leveraging end-to-end training, our method can manage computational complexity while enhancing feature extraction, resulting in a more robust WSI representation. The experimental results on the TCGA and CPTAC datasets demonstrate the superiority of DRE-SLCL in cancer subtyping, cancer recognition, and gene mutation prediction tasks. 
Future work will explore further improvements to the interpretability of the model and extend its application to other types of cancer, aiming to establish a more comprehensive tool for computational pathology.




    

\section*{Acknowledgements}
This work was supported by the National Natural Science Foundation of China (U24A20256, 62473385, and 62202499). We are grateful to the High Performance Computing Center of Central South University for partial support of this work.

{
    \small
    \bibliographystyle{ACM-Reference-Format}
    \balance
    \bibliography{sample-base}
}

\end{document}